# Big Machinery Data Preprocessing Methodology for Data-Driven Models in Prognostics and Health Management

Sergio Cofre-Martel[1], Enrique Lopez Droguett[2], and Mohammad Modarres[1]

[1]Center for Risk and Reliability, University of Maryland, College Park, USA

[2]Civil and Environmental Engineering Department & Garrick Institute for the Risk Sciences, University of California, Los Angeles, USA

**Abstract**

Sensor monitoring networks and advances in big data analytics have guided the reliability engineering landscape to a new era of big machinery data. Low-cost sensors, along with the evolution of the internet of things and industry 4.0, have resulted in rich databases that can be analyzed through prognostics and health management (PHM) frameworks. Several da-ta-driven models (DDMs) have been proposed and applied for diagnostics and prognostics purposes in complex systems. However, many of these models are developed using simulated or experimental data sets, and there is still a knowledge gap for applications in real operating systems. Furthermore, little attention has been given to the required data preprocessing steps compared to the training processes of these DDMs. Up to date, research works do not follow a formal and consistent data preprocessing guideline for PHM applications. This paper presents a comprehensive, step-by-step pipeline for the preprocessing of monitoring data from complex systems aimed for DDMs. The importance of expert knowledge is discussed in the context of data selection and label generation. Two case studies are presented for validation, with the end goal of creating clean data sets with healthy and unhealthy labels that are then used to train machinery health state classifiers.

## 1. Introduction

Cost reduction in sensor monitoring networks and the massive amounts of data generated by them has propelled the application of data-driven models (DDMs) to prognostics and health management (PHM) frameworks. In this regard, most recent research has focused on developing machine learning (ML) and deep learning (DL) techniques for diagnostics and prognostics purposes. At the same time, very few works are dedicated to the study of the data preprocessing steps needed to train these models.

| Acronyms | | | |
|---|---|---|---|
| **Acronym** | **Description** | **Acronym** | **Description** |
| AE | Autoencoder | ML | Machine Learning |
| BNN | Bayesian Neural Network | NN | Neural Network |
| CMC | Copper Mining Crusher | PCA | Principal Component Analysis |
| CNN | Convolutional Neural Network | PHM | Prognostics and Health Management |
| DDM | Data-driven Model | RF | Random Forest |
| DL | Deep Learning | RNN | Recurrent Neural Network |
| DNN | Deep Neural Networks | RUL | Remaining Useful Life |
| HI | Health Index | SVM | Support Vector Machine |
| IoT | Internet of things | VRU | Vapor Recovery Unit |
| KPI | Key Performance Index | | |

Data acquisition and processing are crucial steps in any PHM framework, particularly when DDMs are involved [1]. However, this analysis is frequently left out when comparing or reviewing DDMs applied to PHM [2]. In this paper, we present a preprocessing pipeline for data collected from complex system's sensor monitoring networks for ML and DL applications to PHM.

In PHM, diagnostics and prognostics tasks are commonly addressed through three different approaches: model-based, data-driven, or a combination of these through hybrid algorithms [3]. In complex systems, the availability of physics of degradation models is rare, and if available, these are highly-localized and component-specific [4]. Therefore, implementing model-based approaches is usually not feasible when analyzing complex systems with multiple components and failure modes. Alternatively, DDMs have been widely applied to systems lacking explicit degradation models since these are not system-specific, only relying on robust and reliable data sets. In this regard, significant advances have been achieved in diagnostics and prognostics using DDMs [5]. Nevertheless, published literature has commonly focused on the development of algorithms using benchmark data sets, such as the rolling elements vibrational data from the Machinery Failure Prevention Technology (MFPT) society, the Case Western Reserve University's Bearing data set [6], and the turbofan C-MAPSS data set [7]. These data sets do not accurately represent real-life scenarios since they lack defects commonly observed in data collected from complex systems. Noise contamination, missing data points, feature redundancy, and poor data quality are only a few of the challenges encountered when dealing with data sets obtained from real systems. As such, processing monitoring data from real systems to obtain representative data sets still remains a challenge from a DDM perspective.

Further, the rapid evolution of ML and DL algorithms as well as developments in hardware capacity, have often led researchers to focus on adjusting the models' complexity rather than directly addressing the preprocessing and analysis of the raw data from the system under study. As a result, the importance of the latter is often underestimated, not considering its great impact on the final results [8], [9]. The low number of studies explicitly addressing data preprocessing procedures can be mostly linked to benchmark data sets. Some of these are often provided ready to 'plug-and-play', while others require minimal preprocessing for their use [10]. In the case of PHM frameworks developed for non-benchmarks data sets, the data processing stages converge to system-specific solutions instead of developing standard guidelines. This, despite the common aspects complex systems share when dealing with sensor monitoring data, regardless of the specific system under study. Thus, developing standard preprocessing methods (i.e., cleaning and preparation) for data sets originating from diverse monitoring sensor networks can facilitate the further adoption of ML or DL prognostics and diagnostics models at the industry level.

The main goal of a PHM framework is to provide decision-making tools to optimize reliability, performance, and maintenance policies for systems and their components. Hence, providing tools to preprocess and analyze monitoring data is crucial to close the gap between theoretical models developed in academia and their application in industry. For example, it has been reported that over 2.2B€ are spent yearly repairing unexpected failures in the wind turbine industry [11], while maintenance services can represent up to 35% of the total power generation costs. In the United States alone, remediating corrosion damage in oil and gas pipelines costs over $2 billion [12]. On the other hand, manufacturers are expected to spend over $7 billion per year recalling and renewing defective products [13]. Implementing PHM frameworks to these real systems can be a key step to enhance the optimization of efficient and safe maintenance policies. These are expected to reduce downtime and cost, while increasing the overall availability of the system [14]. In this context, developing robust frameworks for the processing of raw sensor data is essential, as data containing errors or inconsistencies can result in prognostics models with poor performance and, in turn, affect the decision making process [15], [16].

This paper proposes a preprocessing methodology for big machinery data focused on DDMs applied to PHM. As a first step, data cleaning and preparation are discussed and presented in detail. Then, different preprocessing strategies and the effect of additional metadata describing the work conditions of the monitoring system are discussed. It is shown that a successful preprocessing methodology could only be implemented through continuous feedback between data scientists and field engineers. Finally, feature importance analysis and dimensionality reduction algorithms are employed to extract the most informative parts of the data set.

Two case studies from real complex systems are analyzed: a vapor recovery unit (VRU) at an offshore oil production platform and a copper mining crusher (CMC) system. The data preprocessing pipeline can be used as an end-to-end methodology to prepare data from raw sensor readings to create diagnostic models. The pipeline can also be used as a baseline for more complex tasks, such as prognostics and remaining useful life (RUL) estimation [17] or to complement other data types such as vibration signals or acoustic emissions. In both case studies the analysis is focused on mechanical failures, which are expected to have long-term degradation behavior. Hence, this excludes fast developing faults such as the ones found in electrical systems.

The proposed methodology is aimed towards the analysis of data collected from sensor networks data in mechanical systems. These consists of measurands such as temperature, pressure, displacements, and flow. Measurements such as the ones found in electrical systems (e.g., voltage and current), as well as time-frequency signals (e.g., vibrations and acoustic emissions) [18], can also be analyzed as regular sensors. It should be noted that high frequency signals processing techniques for feature extraction are not considered in the proposed pipeline. Similarly, input/output signals commonly encountered in control systems are also not part of the analysis. These types of data are of great importance in complex systems such as urban traffic [19] or structures dependent on controlled systems [20]. However, they rarely present enough variability to be considered in PHM DDMs, and they are not expected to be related to the failure mechanisms of the system.

The main contributions of this work can be summarized as follows:
1. A step-by-step guide for the preprocessing of big machinery data in the context of PHM.
2. A methodology to create health state labels based on the system's operational time and available maintenance logs.
3. Two case studies from different real systems showcasing the proposed preprocessing pipeline.
4. The proposed pipeline is a step towards bridging the gap between theoretical and practical applications of DDMs to PHM.

The remainder of this paper is structured as follows: Section 2 presents and discusses the current literature on machinery data processing and PHM frameworks. Section 3 presents and describes the two case studies used to exemplify the proposed preprocessing pipeline. Section 4 details the proposed data preprocessing procedure for condition-monitoring data in complex systems. Section 5 provides a description of how to generate health state labels. Section 6 illustrates results of ML and DL models. Finally, comments and conclusions are presented in Section 7.

## 2. Preprocessing Sensor Monitoring Data for Prognostics and Health Management Models

Data preprocessing is an essential step of PHM frameworks [21], [22]. Nevertheless, the preprocessing stages are frequently not considered as components that increase the performance and effectiveness of the framework. There are no comprehensive, up-to-date protocols or guidelines on how to handle, organize, and prepare sensor monitoring data to train and implement DDMs in PHM frameworks. Even among reliability researchers, crucial information regarding methods for collecting, handling, and processing data is scarce [23]. Consequently, it is common to encounter published literature that does not give a detailed description of the data preprocessing before training PHM models [24], [25]. This knowledge gap can effectively hinder the applications of DDMs to real systems. This calls for the detailed development of procedures for processing raw sensor monitoring data for using DDM-PHM frameworks in industry settings.

*2.1. Big machinery data*

Big data is often referred to as databases of great size. A more accurate characterization can include the description of tools and methods required for managing a database, determining the computational power needed to store and process the data. Whether to consider a data set as big data or not will depend on the stored data and the desired applications, not just its size in gigabytes. One common concept used to categorize a data set as big data is the 3 V's: Volume, Variety, and Velocity [26]. Volume refers to the size of the data set in terms of memory and the contained number of data points. For instance, processing one terabyte of audio files is not the same as processing a data set of videos of the same size in terms of the contained data points. Variety indicates how heterogeneous the data is when looking at its variables and structure. Combining different types of stored data (e.g., audio and video on the same data set) will increase the Variety of the data set given its lack of structure. The last V refers to Velocity, denoting the data acquisition frequency and the speed at which the information needs to be processed.

Developments and cost reductions in sensor technologies have allowed complex engineering systems to be highly sensorized. That is, systems are equipped with sensor monitoring networks that oversee the system's operational conditions in real time. These networks can record data logs at high frequencies for numerous sensors simultaneously. The constant acquisition of the system's operational data results in data sets of immense volumes and complexity. This brings advantages and drawbacks to the table. On the one hand, if monitored correctly, the collected data is expected to contain information on the system's state of health. On the other hand, having great volumes of complex data is of no use if there are no adequate methods to process it, especially if there is no clear task for what the data can be used for.

Processing data sets collected from monitoring systems can hence be considered as a big data problem. Multiple sensors strategically located at different components of the system result in data with high variety. Sensor networks with high sampling frequencies, like the ones studied in this paper, collect thousands of data points in short periods of time. These networks can also contain dozens or hundreds of sensor features. Hence, we refer to these data sets collected from sensor monitoring data as Big Machinery Data.

*2.2. Processing data for DDM applied to PHM*

Prognostics and health management seeks to provide frameworks for analyzing condition-monitoring data from an end-to-end perspective. PHM is commonly divided into sequential stages, as described in Figure 1 [5]. These stages encompass everything from the data collection process through the sensor network to the system's maintenance policy optimization. Regardless of the system and available data, all PHM frameworks have a diagnostics and prognostics stage. Here, DDMs have gained significant research attention in the last decade since physics-based models are generally not available to describe the degradation processes occurring in complex systems [3]. In this regard, several ML and DL architectures have been applied to a variety of case studies, aiming to extract the maximum amount of information from the available big machinery data. The data preprocessing stage is essential in constructing reliable ML and DL models, normally encompassing greater time and effort than training the models themselves [27].

Preprocessing data involves various steps before training a model, from selecting features that best represent the system under study to normalizing feature values. Depending on the available data and the selected DDM, different preprocessing steps might be needed. Thus, the data curation and preparation processes are often seen as a case-by-case scenario rather than a defined methodology. For diagnostic tasks, a common source of measurements are temporal signals in the form of vibration or acoustic emissions. These signals have traditionally required manual feature crafting based on expert knowledge depending on the signal, resulting in loss of information and time consumption. The extracted features are then used to train regression or classification models, whose performance strongly depends on the features' representability of the system. To this end, particular signal and data processing techniques have been proposed to reduce information loss and improve models' performance [28]. As an example, the capabilities of convolutional neural networks (CNNs) [29] to extract abstract features from raw data have proved to yield better performance than models trained on manually crafted features in diagnosis classification tasks [30]. As such, vibration signals can be transformed into spectrograms or scalograms to obtain a diagnostics model through CNNs [31].

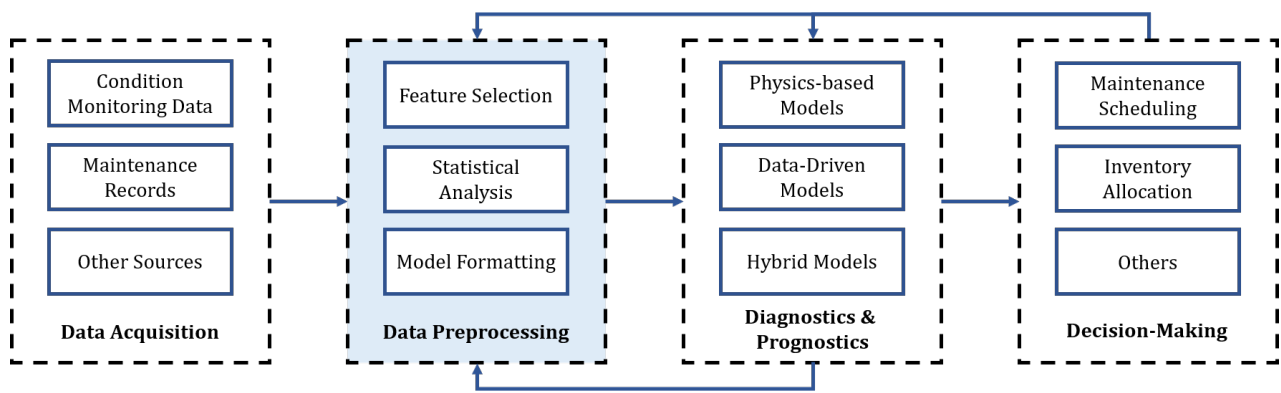

Figure 1: Example of the main stages in a PHM Framework.

Many regression and classification algorithms do not require a specific format for their input data when analyzing sensor monitoring data. Examples of these are support vector machines (SVMs) [32], random forest (RF) [33], and neural networks (NN) [34], where only one value per feature and a label are needed. The latter is required when training supervised models, while unsupervised models only require an input value. In sequential analysis, models such as recurrent neural networks (RNN) require sensor values inputs to be formatted as time windows; however, the sensor values themselves do not need special treatment [35]. Therefore, these algorithms can be trained for RUL estimation or other tasks without many complex preprocessing requirements [17], [36]. This can be observed for ML and DL models developed for fault, error, and anomaly detection [37]–[40]. Likewise, CNNs can also be implemented for prognostics tasks, in which RUL estimation is performed based on time windows created from sensor monitoring data, similar to the ones used for RNNs [41].

Among the published literature describing the preprocessing steps taken to prepare big machinery data for DDM-PHM frameworks, there tend to be important differences in the implemented methods. This occurs even when applying similar techniques to the same case study or data set [42], [43]. Although this variety of preprocessing approaches for the same data is not inappropriate, these examples show that there is no consensus within the reliability community on how sensor monitoring data should be treated for training DDMs. Discrepancies on the preprocessing methodology become an issue when trying to compare or replicate results from different techniques applied to the same data set. The most common examples of this phenomenon are the results reported for the C-MAPSS data set [7]. This benchmark data set has been repeatedly analyzed through multiple preprocessing techniques and DDMs reporting significantly varying results between articles [32], [41], [42], [44]. This could imply that a proposed ML model or DL architecture might present better results due to a more consistent preprocessing methodology than to the models' architecture. Another possible scenario is that the improved performance on a new DDM is due to an improper preprocessing methodology. For instance, the cross-contamination of the training, validation, and test sets, resulting in misleading results due to the overfitting of the models.

*2.3. Differences in preprocessing sensor monitoring data from real and benchmark data sets.*

Sensor monitoring data from experimental setups or simulations have characteristics that are not realistic when compared to data from real systems. In the former, data acquisition frequencies can be coordinated to register sensor logs simultaneously. Sensor layouts can be optimized to enhance the quality of the collected data, where its volume solely depends on the available computational power or time to run the experiment and the sampling frequency. The number of observed failures will also depend on the available time to run simulations. Experimental setups are usually isolated from other elements, while simulating the degradation effects caused by interactions between components is challenging. Noise is commonly added as a Gaussian variable, which does not consider external factors nor the presence of other components. These ideal conditions are rarely seen in data collected from real systems, where sensor readings originate from monitoring multi-component systems, each with its internal physical processes. Here, interactions between sensors can create a high level of noise as well as highly correlated variables. Failure events can be rare for some components and therefore, collecting representative degraded data is unfeasible. Further, as discussed in this paper, expert knowledge is needed to analyze and process the data.

Measured physical variables have a temporal component that needs to be considered. In this context, preprocessing techniques based on statistical analysis, such as the ones utilized in data mining [45], cannot be directly implemented to these data sets. This, given the need to incorporate expert knowledge to ensure that the data represents the system. As an example, resampling among sensor variables can create inconsistencies within the data set. Indeed, components operate following known laws of physics, e.g., heat exchangers are based on the laws of thermodynamics for heat and mass transfer. Resampling can create sensor log sequences that do not follow these laws. Individually resampling each sensor feature would contaminate the data set, thus affecting the performance of a DDM trained on it. On the other hand, DDMs are also highly dependent on the quality and variety of the collected data. Real-world data sets often originate from diverse sources, in environments that add uncertainty and noise-contamination to the data. These data sources can also have different sampling frequencies and sensor or connectivity failures, causing missing values to appear in the collected data set. As such, these non-structured, multi-modal, and heterogeneous data sets cannot directly be used to train DL models, requiring preprocessing and expert knowledge to be analyzed.

Another important difference between benchmark data sets and real systems is the existence of maintenance logs. Information on the stoppages of the system, the details on the failure mechanism and failure modes, and the relationship between components and failures can provide valuable information to develop diagnostics and prognostics models. This metadata can help select data to train the models and label it without specific knowledge for all the components.

*2.4. Current challenges in DDM-PHM applied to complex systems*

Data scarcity and data quality are two factors that currently hinder the practical applications of DL models to PHM based on big machinery data [5], [46]. On the one hand, complex DL models are defined by many parameters (normally among the thousands or millions). This causes DL models to be data-hungry since, to avoid an overrepresentation of the data (i.e., overfitting), the number of data entries used to train a DL model needs to be greater than its number of trainable parameters [47]. Techniques such as data augmentation [48] and transfer learning [49] have been studied to address these limitations, however, their applications are normally bound to a specific case study rather than a generalized methodology. For instance, to compensate the lack of accurately labeled data, Li et al. [50] used data augmentation to generate training sets for a DL fault detection model using experimental machinery vibration data. This approach improved the label representation of the faulty data; however, its implementation cannot be directly extrapolated to similar diagnosis tasks based on different case studies.

The evolution of smart sensors, the internet of things (IoT), and industry 4.0 are expected to alleviate the data scarcity and quality problems, given that sensor networks can generate big sets of data points in a short period of time through a coordinated system [51]. Nevertheless, preprocessing and labeling this data for diagnostics and prognostics tasks remains a challenge due to the need for expert knowledge and time to process the information. Furthermore, since complex systems tend to operate for long periods before a failure is observed, a class imbalance between healthy and unhealthy data points is likely to negatively affect the data set's variety. Indeed, new engineering systems need to comply with more demanding safety and security norms, thus becoming more reliable and developing towards degraded states at a lower rate frequency. With few data points related to failures or degraded states, it is difficult to train a DDM with generalizing capabilities that can discern between a system's future healthy or degraded state.

In summary:
- Massive implementation of sensor networks, the IoT, and industry 4.0 have launched a new era of Big Machinery Data from which PHM can take a big advantage from;
- Few studies focus on the preprocessing of sensor monitoring data for PHM compared to other data sources such as signals and images. This leads to inconsistencies in published results and hinder replicability of works;

- Benchmark data sets have allowed the development of high-performance DDMs, but these do not present the challenges nor the need for extensively preprocessing the data collected from real systems;
- Obtaining high quality data when analyzing real systems with DDMs still remains a challenge. Few failure events lead to unbalanced classes for healthy and degraded states, and accurate labels are still not widely available for complex systems. Thus, expert knowledge is crucial to preprocess big machinery data in the context of PHM frameworks.

## 3. Case Studies

To demonstrate the challenges presented in the preprocessing of real machinery data and the effectiveness of the proposed preprocessing pipeline, we present two case studies from completely unrelated systems. Degraded state labels are created for one component per system, selecting the one that presented the highest number of failures. Namely, the scrubber from the VRU and the crusher from the CMC.

### 3.1. Case Study 1: Vapor Recovery Unit

As a first case study, we analyze a vapor recovery unit, a subsystem at an offshore oil production platform. This process unit is responsible for recompressing hydrocarbons separated from the oil stream during the primary processing stages. The VRU is a complex system that integrates many components such as condensate abatement vessels, heat exchangers, and auxiliary systems for sealing and lubrication. Among these, dry screw compressors with two compression stages can be found. The compressors carry gas from an input pressure of 100 kPag (first stage suction) and 600 kPag (second stage suction) up to approximately 1,900 kPag. The gas is then compressed in the main compressors of the VRU. There are two independent compression trains operating simultaneously in parallel. The system is powered by an electric motor of 13.8 kV and 2.6 MW. The design flow of the first stage is 150,000 Nm³/d and 750,000 Nm³/d for the second stage. Figure 2 shows the layout of the system.

To ensure its correct operation, the VRU unit is equipped with a sensor network comprised of 189 sensors of different kinds such as flowmeters, thermocouples, pressure gauges, and accelerometers. Sensor data logs are registered at a sampling time of 15 seconds from January 1, 2019, until February 18, 2020. For sensor reliability purposes, the sensor network is composed of multiple redundant sensors and thus, many variables are highly correlated. The collected data comes from a real-world operational environment; therefore, it presents noise contamination and missing data logs, among other defects. Along with the sensor readings, a separate file is provided with information regarding the stoppages of the system. Here, for each time the system was stopped, a label and an observation describe the cause of the shutdown. Table 1 shows the possible values for observation and labels. The data set contains a total of 2.4 million points. One of its columns corresponds to Time, while the rest corresponds to the described sensor readings.

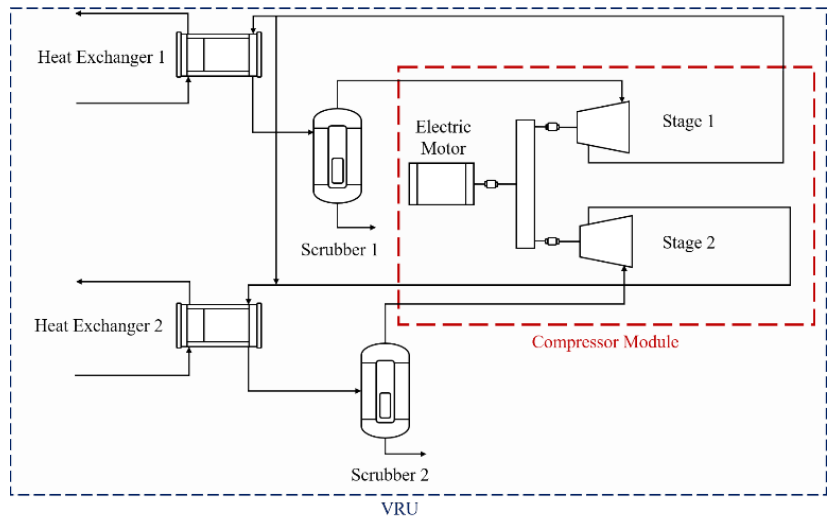

Figure 2. Layout of a vapor recovery unit (VRU).

Table 1. Tag description for VRU stoppages.

| Observation | Labels |
|---|---|
| Undefined, Normal, Stand-by, External, Failure | Undefined, Normal Stop, External cause, Instrument Failure, Mechanical Failure, Electrical Failure, High Suction Pressure, Unidentified Cause |

*3.2. Case Study II: Copper Mining Crusher*

The second case study presented in this paper corresponds to a subsystem of a mining processing line dedicated to copper production: a copper mining crusher. From the collected mineral stockpile, the material is transported through a series of conveyor belts to different components, aiming to make the extraction process of the mineral from the rock easier. Here, two feeders in a parallel configuration transport the raw mineral from a stockpile to a belt conveyor, a sifter, and finally the crusher. The crusher is the principal component in this processing line and the focus of this analysis. A simplified diagram of the monitored components is presented in Figure 3.

A condition-monitoring sensor network collects operational data from the CMC system's components. Sensor measurements, including the time variable, are available from 22 sensors with a sampling time of 2 minutes. Data is acquired from July 1, 2017, to October 1, 2019.

Data logs containing information on all the interruptions and stoppages the system experienced are also provided. The system's logs are available from January 2, 2017, to August 31, 2019. This data documents the interruptions of the system's operation classified into two categories: interruption of normal operation (including programmed maintenance and inspections) and failures detected in the system. There are 2,595 operational interruptions reported during this period, where only 382 explicitly refer to failures in the system.

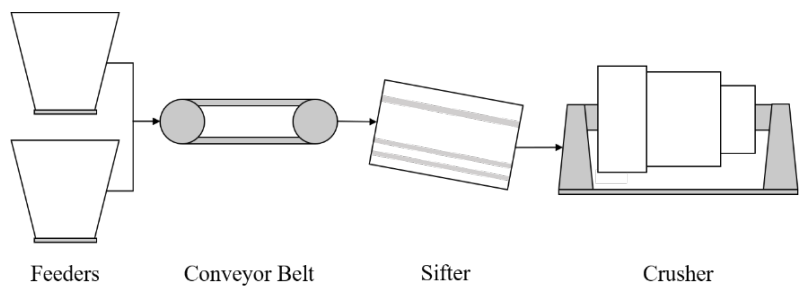

Figure 3. Simplified diagram of copper mining crusher processing line.

Table 2 presents the internal classification of these failures. As can be seen, several causes do not necessarily correspond to failures of the CMC system, such as sensor communication errors, inspection, programmed maintenance, and other external events. Thus, the system detention logs are filtered, only considering events related to failures. The nomenclature used for the available sensor monitoring variables is presented in Table 3.

Table 2. List of most significant interruptions identified.

| Type | Cause |
|---|---|
| **Internal Failure** | Belt failure, Control/Instrumentation failure, Hydraulic failure, Lubrication failure, Mechanical failure, Other Cause, Sifter/Crusher Load Exceeded |
| **External Failure** | Communication failure, Current Exceeded, Electrical failure |
| **Operational Outage** | Cleaning, Inspection, Low Stock Mine, Operation restart, Setting adjustment |

Table 3. Sensor nomenclature in crusher system.

| Sensor Code | Sensor Name | Units |
|---|---|---|
| $I$ | Current | % |
| $P_s$ | Shaft Pressure | kPa |
| $T_r$ | Return Temperature | °C |
| $T_l$ | Socket Liner Temperature | °C |
| $T_f$ | Feeder Temperature | °C |
| $T_e$ | External Temperature | °C |
| $T_s$ | Shaft Temperature | °C |
| $RV_i$ | Ring Vibration (i=1-4) | % |
| $L$ | Vessel Level | % |
| $S$ | Setting | mm |

## 4. Proposed Preprocessing Pipeline

Regardless of the system under study, dealing with big machinery data requires addressing different challenges to ensure its correct processing and interpretation. In the case of machinery data for health assessment, it is common to have dozens or hundreds of sensors, each of which is a potential feature to train a DDM. However, it is important to identify irrelevant features to the system's operation, which are commonly found in real-world data sets. Sensor readings may also present abnormalities such as missing values due to different sample frequencies among sensors, disconnection, malfunction, or external causes, as well as data logs that are not representative of the normal behavior of the system.

Sensor monitoring data come from sensor networks designed to monitor specific measurands of an asset. Therefore, these networks are rarely thought of or optimized to obtain information on the system's state of health. Consequently, complex systems will likely have sensors that provide redundant information, i.e., sensor readings with highly correlated data. Having multiple redundant sensors can be desirable from a sensor reliability perspective, preventing data loss in the case of a sensor malfunction, failure, or replacement. However, when considering sensor values as features for statistical models such as deep neural networks (DNNs), redundancy among the features can subject models to bias. The number of features fed to a DDM will also determine the number of parameters that need to be trained. A lesser number of parameters implies shorter training and evaluation times. Thus, feature selection and reduction are necessary to identify how sensors are related to each other, discarding highly correlated ones.

We present a methodology to preprocess and prepare data sets from big machinery data to train ML and DL models. The proposed methodology is based on statistical analysis and expert knowledge from field reliability engineers. Figure 4 summarizes the proposed preprocessing pipeline, composed of four stages: expert knowledge-based feature selection, statistical analysis, data set preparation, and health state label generation. The data set preparation stage refers to the different data manipulations required depending on the selected DDM and the specific task (e.g., classification, regression). This section gives insights into the first three stages for both case studies presented in Section 3. The last stage of the framework is presented in Section 5.

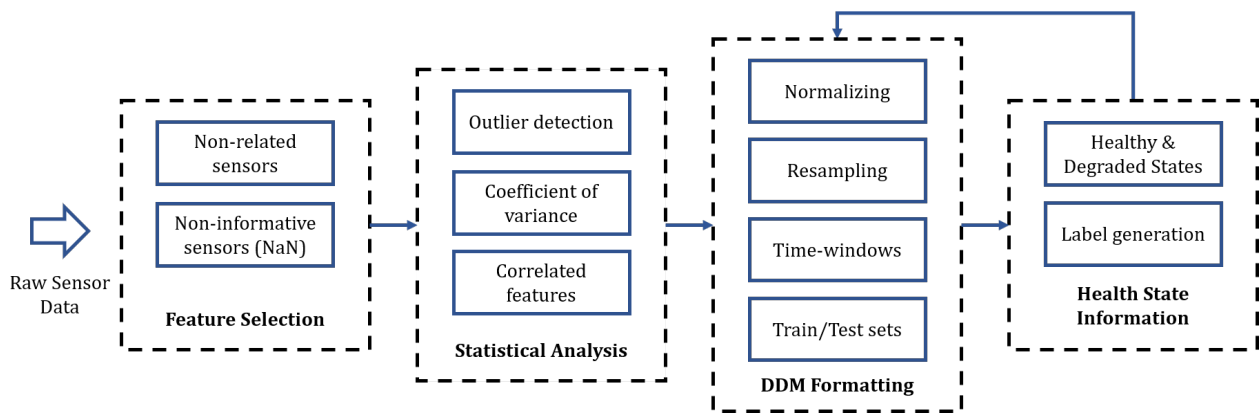

Figure 4. Proposed preprocessing pipeline for sensor monitoring data.

*4.1. Expert knowledge-based feature and data selection*

As part of the feature selection and reduction process, the preprocessing framework identifies and discards irrelevant features and data logs through two different methods. First, expert knowledge from field engineers is used to identify and remove variables unrelated to the system's operation from the data set. Secondly, variables presenting a high percentage of missing or non-numerical numbers (i.e., NaN) are discarded. Identifying sensors that are not related to the system cannot be achieved by a statistical analysis of the variables but rather by an understanding of the system.

As an example, for the VRU system, field engineers assessed that 17 out of the 189 sensors registered in their data base do not belong to the system's sensor monitoring network but rather to external components that are unrelated to it. These sensors are not expected to contain information on the system's state of health and, thus, are immediately discarded. There are also 17 sensors that measure quantities controlled by the operators, such as scrubber levels, oil reservoir levels, and external valve openings. These are also dismissed since they are not related to the system's physics of degradation and their values tend to have low variability.

In the case of the CMC system, engineers reported that the system's components should be analyzed separately since the provided sensor readings are not linked from one component to the other (see Figure 3). This is due to the physical distance between components, and that sensors are placed to measure the components themselves rather than a continuous working fluid. In addition, the main component of the CMC is the crusher, which accounts for 14 out of the 22 available sensors. Therefore, only these sensors are selected and considered in the analysis.

Following this, sensors with numerous missing values need to be identified. If a sensor is missing a high percentage of data points in time, it is considered a non-informative feature. Figure 5a illustrates the original distribution of NaN values for the columns (above) and rows (below) of the VRU data set. Based on the original distribution of missing values, variables presenting 50% or more missing data entries or non-numerical values (i.e., NaN values) are discarded. Using these criteria, 27 uninformative sensors are removed from the VRU data set, resulting in the NaN distribution in Figure 5b. In the CMC, most NaNs values come from sensor communication errors. These values are removed from the analysis by excluding columns and rows with more than 50% and 20% of void entries, respectively. This reduces the void entries to under 2% in this data set.

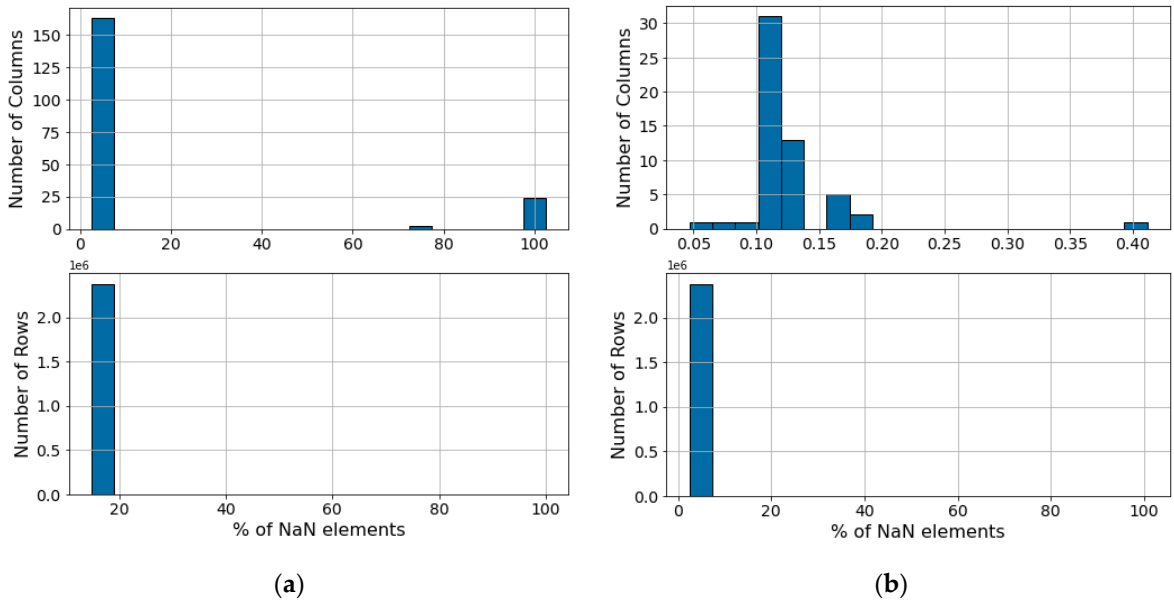

**Figure 5.** VRU NaN distribution for rows and columns: (**a**) Original distributions; (**b**) distribution after removing unrelated and uninformative features.

The temporal nature of sensor monitoring data is another element to be considered. An accurate health assessment model can only be obtained if the collected data is representative of the system. This implies that if there are periods where the system was under operation but not in a representative state, then these values should not be considered in the analysis. Maintenance data logs will usually have information on when and why a system was stopped from its normal operation. Nevertheless, it would be a mistake to assume that every time the system was under operation or when data was being recorded, the system was under normal operation. In this context, the VRU system went through a trial phase for the first five months from the start of the data recording. Hence, engineers suggested that all data logs before May 1, 2019, to be removed before processing the data.

Further, Figure 6 illustrates how maintenance log labels broadly separated data entries into three categories: normal, pause, fault. Pauses are defined as those periods of time when the system is not operating under normal operational conditions. These could be due to scheduled maintenance, low loading conditions, and failure of an external component. Hence, it is likely that these states produce sensor readings that are not representative of the healthy state of the system but do not represent failures or malfunctions either. Although it can be counterintuitive, sensors reading are always recorded for all three categories. Thus, data labeled as pauses should not be considered when preparing a data set to train a diagnostic or prognostic model. This is further discussed in Section 5 when defining the methodology to create health state labels.

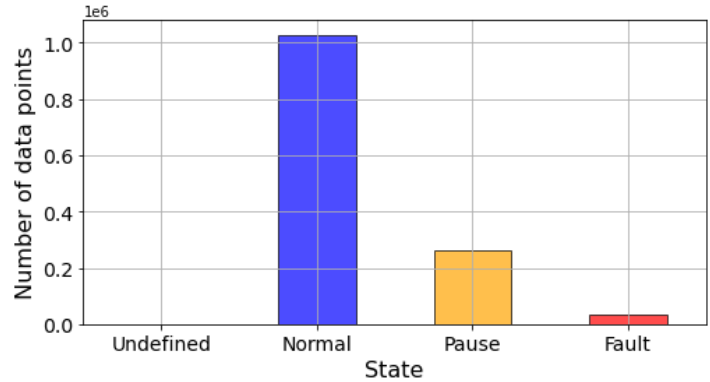

Figure 6. System state distribution based on tag descriptions.

*4.2. Statistical analysis*

After the preliminary feature selection, statistical analysis is used to reduce further the dimensionality of the data set and eliminate redundancies.

4.2.1. Outlier detection

A common practice before applying ML models is to visualize the data to comprehend the behavior of the available features. Visualization also helps to find outliers, which are data points not representative of a variable. The importance of outliers in diagnosis and prognosis models has been studied by Marti-Puig et al. [11] in the context of PHM application to wind turbines. Here, obtaining robust and accurate diagnostics and prognostics models is crucial due to the competitiveness that renewable energies need to present as an alternative to fossil fuel. In their study, the authors conclude that a systematic outlier detection leads to discarding points which could have been representative of the normal operation of the system. Their study shows how the presence of outlier points does not necessarily mean that the system is presenting an anomalous behavior.

From a ML and DL perspective, detecting and removing outlier points can improve the representativeness of the models for reasons similar to the ones presented for the feature selection. However, over-removing data points can lead to overfitting models with low generalization capabilities. This leads to poor performances from the models when evaluated in new unseen data, particularly when faulty states are observed less frequently. Sometimes, what may seem to be an outlier point during the operation of the system might just be the presence of less frequent operational modes, as well as incipient faulty states. It is important to consider that although a complex system will be affected by external noise, the processes that are being measured by the sensors will still follow the laws of physics, presenting their natural inertia.

A common practice for outlier treatment is to assume a Gaussian behavior of the data and remove points that lie outside a determined confidence interval (normally 95th percentile) [22]. The same assumption is made when using the Mahalanobis distance to set a threshold for outlier removal [52]. However, these approaches completely ignore the temporal nature of the data. Hence, a better approach should consider both the temporal behavior of the measurements and statistical metrics to find and discard outlier points. This approach is more time consuming since each sensor variable needs to be analyzed individually. It also requires determining the thresholds for points to be considered as outliers, and thus engineering knowledge can bring important insights for this decision. Complementing both statistical and temporal metrics, the resulting data set is more representative of the actual system's operation, which will result in a more robust model.

Figure 7 showcases examples of how detecting and removing outlier points can improve the quality of data from a VRU accelerometer and a crusher barometer, respectively. In the case of the VRU, Figure 7a shows the temporal behavior of an accelerometer before (above) and after (below) the outlier detection process. Figure 8a also shows the corresponding box plot at a 95% confidence interval. It can be observed that if only a statistical analysis were considered for the outlier detection (i.e., dots on Figure 8a), normal operational values would be discarded even though they do not correspond to outliers. For instance, Figure 8a suggests that any value above 0.25 g should be considered as an outlier. However, clearly shows that these values are part of the system's normal operation. Considering both the temporal behavior of the data and its box plot, 0.3 g and 0.1 g are used as cutoff points for the upper and lower bound of the sensor, respectively. As shown in Figure 8a (bottom), the boxplot still suggests the removal of outlier in both bounds. Nevertheless, the temporal behavior of the signal does not show relevant outliers (Figure 7a).

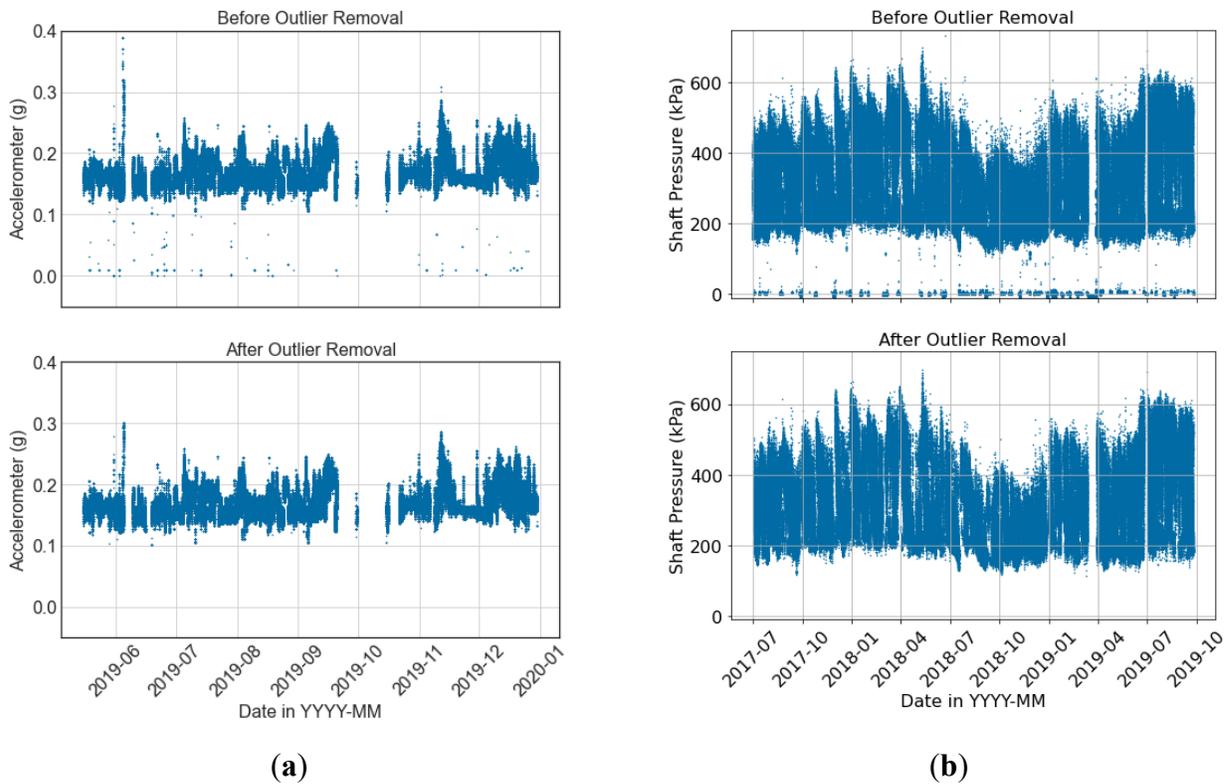

Figure 7. Sensor values in time before (top) and after (bottom) outlier detection: (a) VRU accelerometer; (b) Crusher shaft pressure.

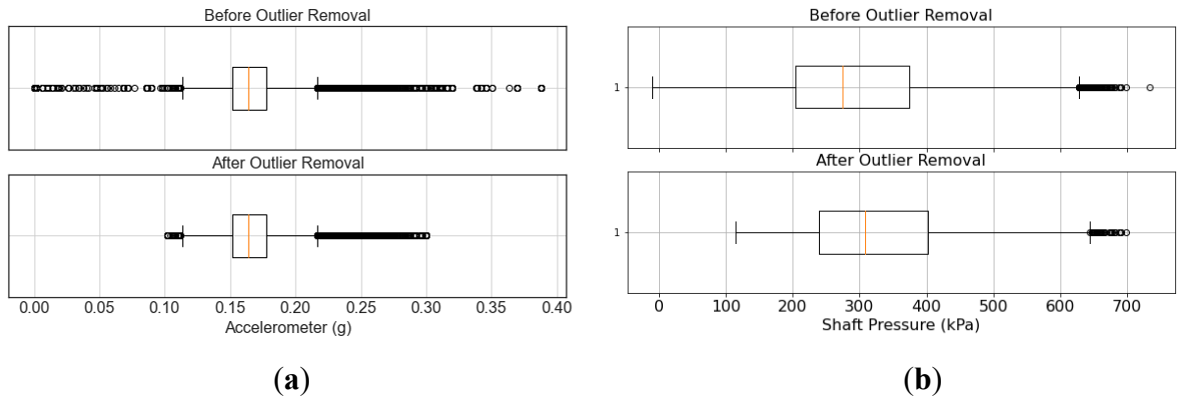

Figure 8. Boxplots before (top) and after (bottom) outlier detection: (a) VRU accelerometer; (b) Crusher shaft pressure.

Figure 8b presents a boxplot for the shaft pressure of the crusher. On the one hand, it can be observed that the statistical analysis suggests that there are no outlier points on the lower bound. However, the temporal behavior shown in Figure 7b clearly shows several outliers at low values. On the other hand, points on the upper bound of Figure 7b are within the nominal operational conditions, and there are no anomalous values. Nevertheless, Figure 8b considers all values above 640 kPa as outliers. Considering both temporal and statistical criteria, values below 100 kPa are considered outliers, while no outliers are removed on the upper bound of the shaft pressure. Before the outlier removal, the shaft pressure presents a mean and standard deviation of 274.98 and 142.96 kPa, respectively. Removing outliers results in a mean pressure of 325.09 kPa with a standard deviation of 104.52 kPa. This is an example where the outlier removal is completely the opposite of what the statistical analysis suggests, emphasizing the importance of incorporating the temporal visualization to the statistical analysis, and the consideration of expert knowledge in the form of nominal operational conditions. Further, reducing the standard deviation can have a great impact when training deep learning models, given that input features typically need to be normalized before being fed to the model during training and evaluation.

Since a DDM diagnostic model is expected to differentiate healthy states from abnormal and faulty states, the outlier detection methodology is applied only to the data labeled as healthy. This approach will also make the model's predictions more conservative and risk-averse since it is less likely for the model to yield false negative values, i.e., classify a degraded state as healthy. This outlier detection and removal process is applied to all the remaining sensor variables in the data set after the feature selection.

### 4.2.2. Feature reduction

The dimensionality of the data sets can be further reduced by discarding redundant variables as well as features with low variability. This reduces the bias of the trained DDM and the input noise contamination. Two criteria are used to select features based on statistical analysis: the variables' individual temporal variation and their correlation. Features with low variability do not add meaningful information to the model since they have a relatively constant value, and thus they are not useful for a diagnostic DDM. The coefficient variation ($cv$) is considered for this purpose, which is defined as the ratio of the standard deviation to the mean of a feature:

$$cv = \frac{\sigma}{\mu}, \tag{1}$$

where $\sigma$ and $\mu$ are the standard deviation and the mean of the feature, respectively. Considering data labeled as healthy, sensor features with a $cv < 0.05$ are discarded. In other words, if the variance is less than 5% of the mean, the feature can be considered as random noise pollution [53].

The second criterion considers the correlation among variables (i.e., redundancy). Pearson's correlation is computed among all sensors. The presented case studies showed that sensors with a Pearson's correlation higher than 0.9 are in physically proximity, and normally measuring the same variable (e.g., temperature). On the other hand, correlation values under 0.8 correspond to sensors that were either further apart in the system or measuring unrelated variables of the system. As such, for sensors presenting a Pearson's correlation higher than 0.95, one sensor is discarded, and the other is kept. This selection is made randomly since the information contained in the features is considered to be equivalent. The selected threshold for the coefficient variation and correlation will depend on the system under study and the available data.

For the VRU case study, the statistical analysis resulted in removing another 62 sensors, leaving 66 informative condition monitoring variables from the original 188. For the CMC, only 14 sensors are selected from the 22 related to the crusher. This implies that the total number of features was reduced to 35.1% and 50% of the original size for the VRU and CMC, respectively. Note that feature reduction through statistical analysis and outlier detection can be an iterative process. For example, one might want to remove outliers again after highly correlated features have been detected. Another option is to remove the correlated variables and then remove the outliers for the first time (all over but without the selected variables).

*4.3. Further processing*

Each complex system and the data collected from its monitoring sensor network will have different elements and challenges. So far, we have covered the common minimal preprocessing steps that should be applied to any big machinery data set. However, the following are additional challenges that can be encountered in the preprocessing of data sets from real systems.

4.3.1. Categorical and text variables

DL algorithms used in PHM are generally not suitable to process text data and sensor monitoring data simultaneously. If the text variable is categorical (e.g., 'open', 'closed'), the text data can be transformed into a one-hot encoder variable or numerical values (e.g., 0 and 1). If the text comes from a maintenance log, it likely contains descriptions and information that can be processed through text recognition algorithms such as natural language processing (NLP) [54]. This can later be used as a complement to the main DDM-PHM framework. It is important to note that the statistical analysis presented in Section 4.2 would not be suitable for categorical variables.

4.3.2. High dimensionality and encoding

Although we have shown how feature selection through statistical and redundancy analysis reduces the dimensionality of the data set, further reduction may be needed. The available volumes of data and the chosen algorithm will constrain the number of features used to train a ML or DL model. This is also determined by the available computational processing power to process high-dimensional data. To this end, tools such as principal component analysis (PCA) [55] and autoencoders (AE) [56] could be used to reduce the data into the desired dimension. This comes at the cost of information loss, but to a lesser degree than discarding additional features. These techniques could also allow to visualize the data for clustering and classification purposes [57].

4.3.3. Normalization

Normalizing the input features to ML and DL algorithms produces models with better performance, faster and stable training, and prevents bias toward any particular feature [29]. Two common approaches are the Min-Max scaler and the Standard scaler defined by Equations 2 and 3, respectively,

$$X_{minmax} = \frac{X - X_m}{X_M - X_m}, \qquad (2)$$

$$X_{stand} = \frac{X - \mu}{\sigma}, \qquad (3)$$

where $X_M$ and $X_m$ are the maximum and minimum values of each feature, while $\mu$ and $\sigma$ are its mean and standard deviation, respectively. Both approaches scale the data through a linear relationship, with the main difference being that the Min-Max scaler yields values between 0 and 1, while the Standard scaler provides values that are normally distributed. There is no rule of thumb on which scaler should be used. The chosen approach will greatly depend on the selected algorithms as well as the values of the data. If feature values range in the positive domain, using a Min-Max scaler will keep the positivity of the data. This can ease the analysis of the variables and the model training process. On the other hand, the standard scaler will not restrict the values to a certain range, which is desirable if the feature presents high variability and a wide domain. Categorical variables should not be normalized. The normalization process should only be applied after the data is separated into training, validation, and test sets to avoid data cross contamination, as discussed in Section 6.1. The scaler is then fitted to the training set, and then the validation and test sets are transformed based on the information from the training set.

4.3.4. Missing values and resampling.

As discussed in Section 4.1, sensor monitoring data tend to present missing values due to sensor malfunction, human error, and difference in sampling frequency, among other causes. Once the feature selection is performed, it is still possible to find non-numerical values remaining in the data set, even though uninformative features have been removed. If the missing values account for a small percentage of the data in the time domain (i.e., rows) then these rows can be discarded. In machinery data, unlike other problems where ML and DL models are implemented, the temporal nature of the data collection method results in great volumes of data to be registered, and these missing values might not represent an important percentage of the remaining data. This is the case for the VRU and CMC, where less than 2% of the remaining rows present missing values. Discarding rows of data with missing values will not affect most ML and DL models' training process. However, some models require continuous sequences of data, such as RNNs [58]. In this case, a more in-depth analysis must be performed since time windows with missing values will hinder the performance of the trained model.

The most simplistic approach to deal with missing values is to perform interpolation. Linear interpolation can be useful only if few data points are missing for each sensor feature and are not continuously missing for an extended period. More sophisticated interpolations can be implemented using polynomial functions or Bayesian interpolation [59], as well as fitting parametric functions [60]. DL models have also been used to complete the missing values with predictions from a RNN regression of the time-series [61]. Another approach is to fill the missing values with a value of zero instead of using an average or resampling method, thus creating a "natural" dropout of the input layer in a DL model [62]. This adds uncertainty to the model, but in turn, makes it more robust to noise and, evidently, missing values.

If missing values cause the loss of valuable data, data augmentation is an alternative to generate new synthetic data with labels based on the original samples. Traditionally, data augmentation techniques have been used for computer vision (i.e., image processing) in classification and NLP tasks. These techniques cannot be directly applied to multivariate time series due to the possible complex interaction among the features and their dynamic behavior [59]. Thus, blindly applying data augmentation techniques would result in synthetic data that is not representative of the system under study. Time series data augmentation deals with the limited data available for certain classes, taking into consideration the temporal dependency of the data. Data sets for classification, anomaly detection, or regression require different data augmentation treatments.

Although it would be impractical to list all the possibilities of missing value treatment, they all have one thing in common: none consider the physics laws for multivariate series. This can become a problem when studying the physics of degradation based on the data. Resampling, interpolating, and augmenting data for multiple features at the same timestamp will likely result in feature combinations that might defy the laws of physics, and therefore the system under study.

For both our case studies, we chose not to use resampling or other techniques to fill missing values with synthetic ones since the latter may not follow the system's physics of degradation and might impact the performance of the models.

## 5. Generating health state labels

Depending on the required analysis, different DDMs can be trained on sensor monitoring data that has been preprocessed using the proposed methodology. In the context of PHM frameworks, the two most common tasks are diagnostics and prognostics. Regardless of the chosen approach, most ML and DL applications to PHM have some degree of supervised learning, thus requiring accurate labels for the models' training process. Many benchmark data sets provide labels from the offset, while in others, their creation is straightforward. This is not the case for big machinery data sets, where multiple components have multiple failure modes and infrequent failure events. This section presents a methodology to create labels for diagnostics (i.e., classification) tasks.

*5.1. Failure mode selection*

Generating a classification data set with healthy state and failure mode labels can be challenging when there is no knowledge of the system's physics of degradation. Multiple failure modes can also hinder the label generation when accurate information on what caused the system's failures is not available. It is here where including expert knowledge on the system can be crucial to analyze machinery data from real systems. If maintenance logs are available, an analysis of the possible failure modes of the components can be made.

In the case of the VRU, only the failed component's name is reported in the logs, but not their failure modes, whereas in the CMC system failure logs describe several failure modes for each component. Figure 9 illustrates the failure frequency for each component of the VRU system. There are 19 failures in the elapsed time-period, and most failures are related to instrumental and scrubber failures. Given that instruments are not considered part of the system, the scrubber is the component that presents the highest failure frequency.

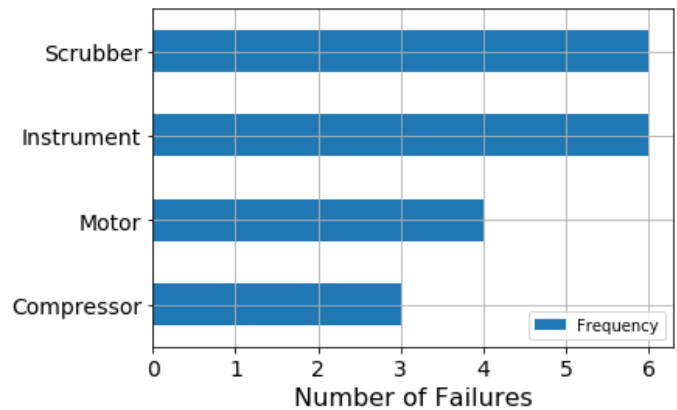

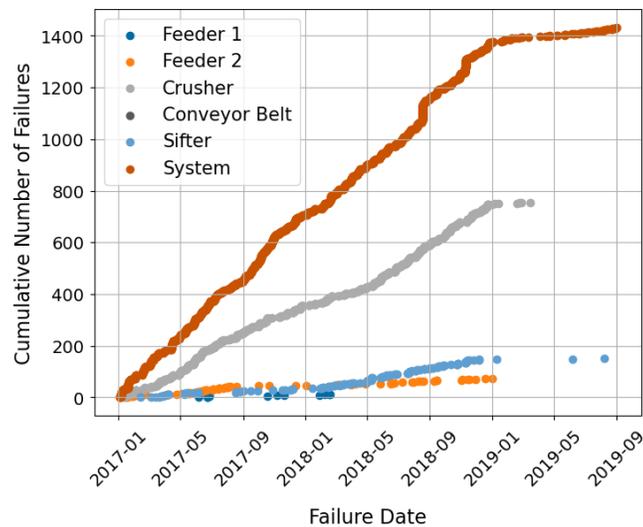

Regarding the CMC system, Figure 10 shows that the crusher is not only is the main component of the system, but also the one that presents the greatest number of failures. Figure 11 shows a breakdown of the possible failure modes of the Crusher. Here, the mechanical and hydraulic failure modes are the most relevant to the system.

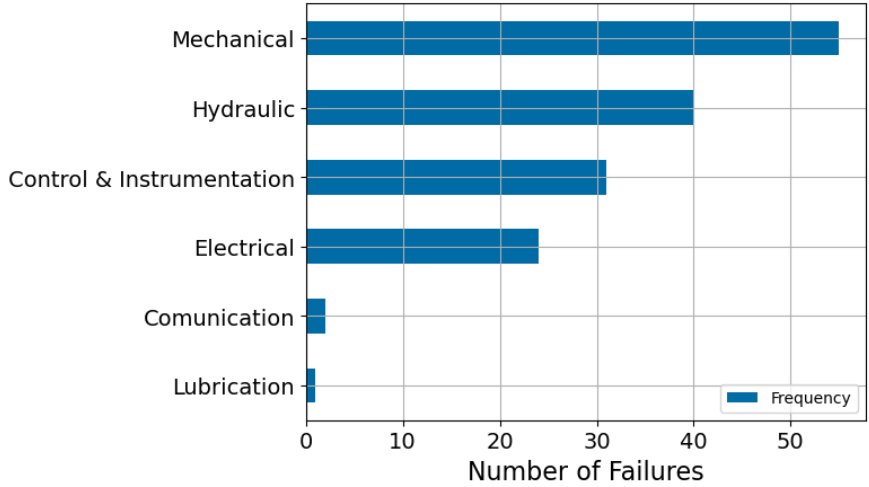

*5.2. Degraded labels*

A common practice to generate health state labels for diagnostic tasks is to use Health Indexes (HIs) or Key Performance Indexes (KPIs) to define thresholds and separate the system's states of health [17]. If there are no HIs or KPIs associated with the system, having accurate diagnostics labels associated with the sensor monitoring variables becomes a challenge. In complex systems it is rare to have these indexes for each of the components or the system. In turn, failure and maintenance logs are likely to be registered regardless of the system, providing quality information on how the healthy state of the assets may look like.

By assuming that a component must be at a degraded state before a failure event, the system's degraded states can be defined as the period immediately before each failure event. Then, operational times that are not close to any failure event or stoppage can be considered as a healthy state. A formal definition for this label generation is the following. Let us consider a system $S$ composed of $N$ components. Each component is denoted as $C_i$ with $i = \{1, \dots, N\}$. $C_i$ has $m_i$ sensors associated to it, where each sensor is denoted as $x_k^i$ with $k = \{1, \dots, m_i\}$. For each component $C_i$ there is also a pool of possible failure modes $FM_j^i$ with $j = \{1, \dots, n_i\}$, where $n_i$ is the number of possible failure modes for $C_i$. For simplicity, it is assumed that all failure modes are independent. The nature of the systems under study allows us to assume that components do not present common cause failures. Now, let us denote $t_i^j$ as the time the system was stopped due to the failure mode $j$ from component $C_i$, and $t_S^p$ as the time when the system is stopped or paused for any reason that is not one of the known failure modes (e.g., external causes or pauses).

When a component $C_i$ fails at $t_i^j$ due to a failure mode $FM_j^i$, it is safe to assume that there is a period preceding the failure event where the component is at a degraded state associated with failure mode $j$. It can also be assumed that the progression of this degradation led to the failure event. As such, we define a time window $\Delta t_w$ before $t_i^j$ where the system is at a degraded health state. That is, the degradation of the component starts at $t_i^d = t_i^j - \Delta t_w$ and it ends at $t_i^j$, corresponding to the failure event registered in the maintenance log. All data points that fall into a time window $\Delta t_w$ are labeled as '*degraded $FM_j^i$*', where $FM_j^i$ is the failure mode associated to the degradation.

Since degradation processes are dynamic, the transition from a healthy to a degraded state does not happen from one sensor timestamp to the next. Instead, given the inertia of the process, there should be a transition time $\Delta t_{tr}$ between the end of the healthy state and the beginning of the degraded state. This serves a double purpose, where the system is allowed to fully progress into a degraded state, while ensuring no cross-contamination between the data labeled as degraded and healthy, respectively. The start of the transition state is then defined as $t_i^t = t_i^d - \Delta t_{tr}$. Figure 12 illustrates an example of this health state labeling methodology, where suction pressure and temperature points in time are shown for the VRU and CMC systems, respectively. The three defined states (i.e., healthy, transition, and degraded) are color-coded in blue, yellow, and red, respectively.

It is important to note that the chosen time window depends on the component and the failure mode that caused the system's detention. A practical degradation window should be long enough that it will allow engineers to detect the failure and act on it before it occurs. Normally, the earlier a degraded state is diagnosed, the better. This would mean setting the start of degradation days (or even weeks) before the failure event. However, the longer the time before a failure is considered as the degradation initiation point, the more likely it is to mislabel healthy states. Mislabeling healthy states can have a negative impact on the model's performance and is likely to increase false negative and false positive predictions. Transition states should also be considered for a warm-up and cool-down periods, i.e., the periods when a machine is started or turned off after or before a being completely shut down, respectively.

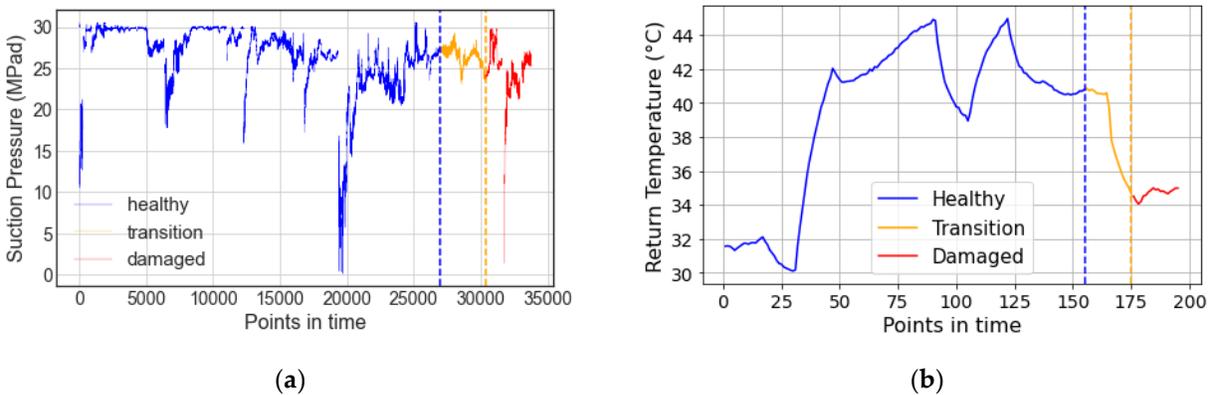

(a)          (b)

**Figure 12.** Example of different designated health states on sensor measurements: (**a**) Suction Pressure (VRU); (**b**) Return Temperature (Crusher).

For simplicity and demonstration purposes, we have considered the same time window for the degradation and transition states for both the scrubber and the crusher from the VRU and the CMC systems, respectively. A time window $\Delta t_w = 2$ hrs was used for the degradation, and a time window $\Delta t_{tr} = 3$ hrs was used for the transition period. A 30 min warm-up and cool-down period was considered for the components and system. The selected degradation time window is enough for engineers to react preventively to the failure events and ensure that all the points contained correspond to the degraded state, while the transition window is long enough to ensure there is no overlapping between healthy and degraded states.

## 6. Results from Machine Learning and Deep learning models

The preprocessing and health state label generation methodologies presented in this paper are designed to yield representative and reliable data sets in the context of PHM frameworks. Diagnostic models can be directly trained using the created data sets and labels, both for ML and DL algorithms. This section presents and discusses different results when training ML and DL models using the generated data sets for both case studies. SVM, RF, NN, and Bayesian neural networks (BNNs) models are trained to showcase the effectiveness of the proposed methodologies.

### 6.1. Data set preprocessing

Before training the diagnostics DDMs, both data sets are separately prepared using the preprocessing methodology presented in Figure 4 and described in Sections 4 and 5. First, unrelated and uninformative sensor features are removed. Then, outlier points are individually detected and removed for each of the remaining features based on both statistical and temporal analysis. The processing follows by identifying and removing features with low variability. Similarly, when two or more sensors are found to be highly correlated, only one sensor is randomly kept. Once these features are removed (i.e., unrelated, uninformative, invariant, and correlated) the feature reduction process is finished. Following this, the original raw data from the remaining features is reloaded. This step is important, since the outlier detection process might have eliminated sensor logs based on features that were discarded during the feature selection process.

Maintenance logs are analyzed to create health state labels. Operational time intervals are defined, starting from the end of a stoppage event up until the beginning of the next stoppage, pause, or failure event. In the case of the VRU system, the compressor module has a sensor measuring the motor's phase current usage, where a positive value indicates that the system is under usage. This threshold is used as additional information to ensure the system is operational. For both case studies, all data points that are not labeled as "in operation" are then discarded. This results in operational time-windows, whose end is related to a specific cause (e.g., pause or failure). The time-windows are thus tagged based on the stoppage cause. These tags are later used to define the degradation and healthy labels, as described in Section 5.

Data labeled as healthy are grouped and saved for the training of the classification models. Data labeled as degraded are saved separately based on common failure modes. This allows to use the data to train models per failure modes or for the system as a whole, as needed. Finally, the outlier detection and removal processes are performed on the healthy data set, in order to remove possible anomalous behavior and noisy data. Removing outliers only from the healthy state allows to train diagnostics models that are more conservative, since they are likely to yield less false healthy states. It also allows the trained model to perform as both a fault and anomaly detection model.

*6.2. Training, validation, and test sets*

Training ML and DL models requires separating the data into three different sets: training, validation, and test sets. First, data is split between a train and a test set, the latter typically accounting for 10% to 25% of the whole data set. Next, the train set is further divided into training and validation sets, with the latter usually representing up to 15% of the train set. Finally, the model's hyperparameter and trainable parameters are adjusted by training the model on the training set and assessing its performance on the validation set. The test set is only used to evaluate the final trained version of the model.

When separating the data set, one must ensure that there is no cross-contamination between the resulting subsets. If the train set contains information (e.g., same data points) from the test set, then it is likely that the model will present overfitting. Further, to accurately train and test classification models, it is ideal that all classes have similar sizes, especially during the testing stages [63]. Having balanced classes allows to easily compare classification metrics such as the confusion matrix, precision, recall, and f1-score.

Data sets are separated as follows:
- Data labeled as degraded is split into train and test sets by randomly sampling the degraded-labeled data without replacement. The test and train sizes are 15% and 85%, respectively;

- Since there are more healthy-labeled data points than degraded data points, the healthy data set is sampled without replacement with a sample size equal to the size of the degraded-labeled data sets. This results in a train and test sets with balanced classes;
- 10% of the train set is then randomly sampled to obtain the validation set, while the remaining 90% is used for the model's training.

*6.3. Data-driven models and results*

Four different ML and DL models are presented. SVM and RF models are trained using the scikit-learn Python library [64], while NN and BNN are trained with Tensorflow 2.4 [65]. A 5-fold cross validation is used to find the best hyperparameters in the ML models. In the case of the DL models, stochastic grid search [66] is used to find the best hyperparameters. This is done for both case studies separately.

Classification metrics for all the trained models on both case studies are presented in Table 4. In general, both case studies present encouraging results for at least two of the trained models. RF and NN seem to be the most robust models, while SVM and BNN struggle with the CMC system. Further, Table 4 shows that models perform differently for both case studies, which is a sign of no cross-contamination of the data sets and a good degradation label definition approach. The physical differences between both case studies justify the models' performance differences from one system to the other. Since the VRU's degraded labels are defined based on components' failure times, it is reasonable to expect better results in its models when compared to the CMC's. This idea is further reinforced with the results obtained for the SVM and BNN models, where false healthy predictions are negligible compared to the false degraded labels. Hence, the trained models are conservative. Figure 13 illustrates the evolution of the training and validation loss (loglikelihood) and accuracy during the BNNs' training process on the VRU. No significant overfitting or underfitting is observed. However, by accounting for uncertainty, the model loses precision on its prediction. This hinders the classification performance and the training process, which can be observed in Figure 13. Hence, the BNN struggles to differentiate one specific failure mode from the healthy data and has difficulties converging to an optimal solution.

Table 4. Classification metrics per case study and model with proposed methodology.

| | | Training | Validation | Test | | | |
|---|---|---|---|---|---|---|---|
| **Case Study** | Model | Accuracy | Accuracy | False Healthy | False Degraded | Recall | F1-score |
| VRU Scrubber | SVM | 0.97 | 0.97 | 0.00 | 0.03 | 0.97 | 0.97 |
| | RF | 0.99 | 0.99 | 0.00 | 0.01 | 0.99 | 0.99 |
| | NN | 0.99 | 0.99 | 0.00 | 0.01 | 0.99 | 0.99 |
| | BNN | 0.97 | 0.97 | 0.02 | 0.01 | 0.96 | 0.96 |
| CMC Crusher | SVM | 0.72 | 0.73 | 0.03 | 0.21 | 0.97 | 0.97 |
| | RF | 0.99 | 0.96 | 0.01 | 0.02 | 0.95 | 0.95 |
| | NN | 0.93 | 0.93 | 0.00 | 0.05 | 0.91 | 0.91 |
| | BNN | 0.70 | 0.72 | 0.03 | 0.26 | 0.72 | 0.70 |

Moreover, given how healthy states are defined, a diagnostics classification model is expected to register as degradation both degraded states as well as anomalies. Thus, one would expect the model to be accurate when classifying healthy states, which is desirable from a reliability point of view. This idea is reinforced further when considering the outlier detection and removal process described in Section 4.2.1 (reducing unnecessary anomalies which could have been detected otherwise). Evidence of this is the results obtained for the SVM and BNN models on the CMC system, shown in the difference between false positives and negatives in Table 4.

The high number of false degraded states can be due to the presence of anomalies in the system. It can also be related to how the transition window was defined between the healthy and degraded labels. Indeed, since the same time window is considered for both case studies, and labels are defined on a time base, it is plausible that some degraded and healthy labels overlapped. In this regard, it is expected that the SVM and BNN should have the greater performance drop. SVMs are known for struggling to separate overlapping classes, while BNN rapidly lose precision with contaminated data given the uncertainty in the layer's weights.

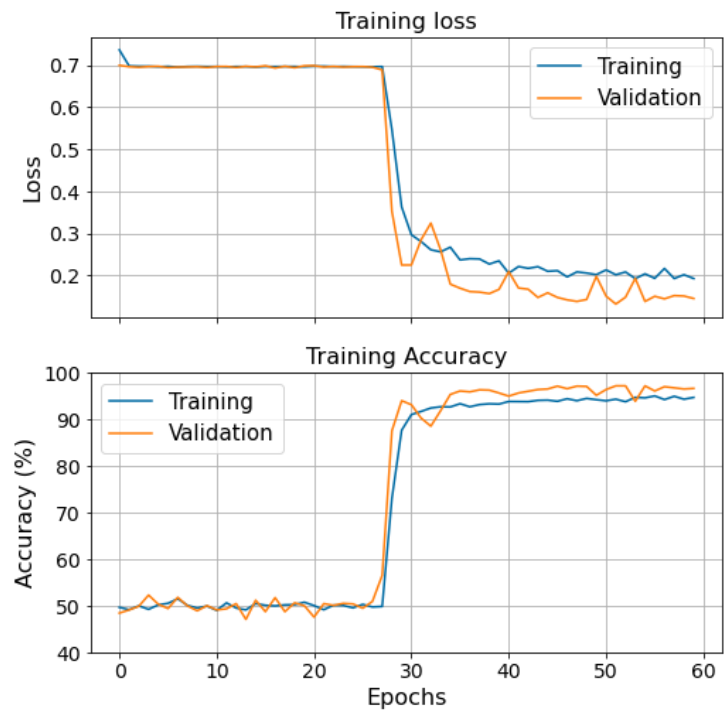

Figure 13. BNN training curves for the VRU Scrubber.

*6.4 Comparison to other preprocessing methodologies*

The proposed framework considers the basics steps that are required to prepare big machinery data for the training of DDMs in the context of PHM frameworks. This considers multiple steps from handling raw sensor data to state of health label generation. One of the most important aspects of the proposed approach is the integration of expert knowledge through the use of maintenance logs and feature selection.

To evaluate the proposed framework's effectiveness, the obtained results are compared to those of diagnostics models trained considering only minimum preprocessing steps. These are: eliminate those variables with constant values in time, replace NaN with zero values, and normalize the feature values. A non-supervised approach is then used to create an anomaly detection model through AE [67], which is evaluated through the reconstruction error of the input values. This AE model is expected to yield high reconstruction errors for anomalous data that are statistically different from the training data. A more elaborate comparison can consider the label generation process based on maintenance records, proposed in Section 5, without a transition state between the healthy and degraded labels (i.e., $t_i^t = 0$). These labels are then used to train supervised models using the data with minimal preprocessing. A third approach would consider dimensionality reduction techniques, as the ones mentioned in Section 4.3.2, which can be achieved through AE or PCA [68]. When applying AE or PCA there is loss of information and interpretability on the original features. Nevertheless, the variables in the encoded space (or principal components) are expected to have a denoised representation of the original input data while retaining essential information from the original space. A reduced dimensionality is also commonly used as a latent space representation that allows visualization, clustering, and metric definition for diagnostic purposes [69]. Hence, the preprocessing methodology proposed in this paper can be used as an alternative to these methodologies, or as a complement in the case the feature reduction process described in Section 4 is not enough.

The following preprocessing scenarios are used to compare the proposed preprocessing methodology:

1. An anomaly detection model through AE is trained on the raw data. The training process considers balanced classes for the test set, but not for the training set. Labels are only considered to evaluate the performance of the model, but not for its training processes. NaN values are replaced with zeros, and input values are normalized using the standard scaler (Equation 3).

2. A PCA is performed on the same raw data as the first scenario for feature dimensionality reduction. A supervised diagnostic model is then trained using principal components as input features, and time-based health state labels defined in Section 5.
3. An AE for dimensionality reduction is performed on the same raw data as the first scenario. A supervised diagnostic model is then trained using the AE's latent space as input features, and time-based health state labels defined in Section 5.
4. Scenarios 2 and 3 are repeated but considering balanced classes for both the training and test sets.

The comparison is performed on the VRU system since it is the most complex of the two case studies presented in this paper. Before training the AE and PCA models, the latent space dimensionality needs to be determined for both the anomaly detection model and the dimensionality reduction process. For deep AEs, one can perform a grid search to find the best latent space dimensionality. However, this is time consuming and escapes the scope of this section. Alternatively, an explained variability analysis can be performed through the PCA, where a threshold is defined for the explainable variance one is willing to give up in exchange for a dimensionality reduction. Hence, for simplicity, we perform an analysis on the cumulative variance from the PCA, setting a threshold of 90% of explainability to select the principal components. Given the similarity between AEs and PCA, the same dimensionality is used for the AE's latent space. Figure 14 illustrates the cumulative explained variance based on the number of principal components. The first 26 components result in a 90.4% of cumulative explained variance, and thus this is the chosen dimensionality for the AE and PCA.

Table 5 presents the results for the four preprocessing scenarios described above. Notice that results are presented only for the RF and NN. This is due to two reasons. Firstly, when considering imbalanced training classes and over a million training samples, neither the SVM nor BNN models converged in a reasonable time (after three days running the training process was stopped). This means that for scenarios 1-3 the training becomes intractable for these models. Secondly, RF and NN are the models with highest performance in Table 4, thus it is natural to focus the comparison on these models.

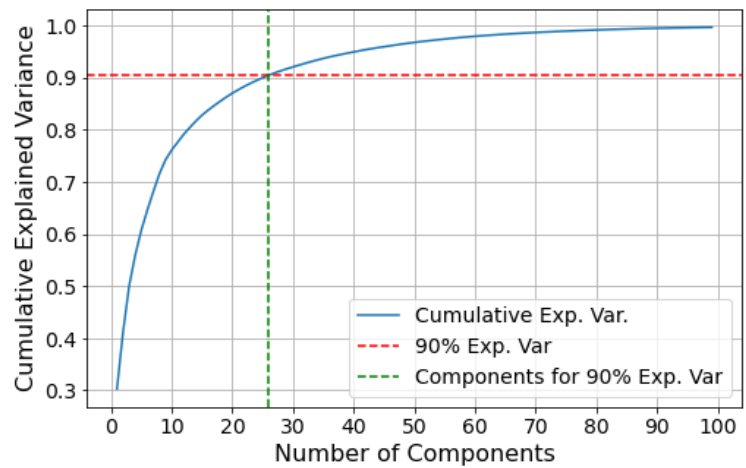

Figure 14: Cumulative Explained Variance per principal component in PCA analysis.

The unsupervised AE used for scenario 1 consists of two layers, one for the encoder (latent space) and a second for the decoder (reconstruction). Rectified linear unit (ReLU) is used as activation function.

Figure 15 shows the reconstruction error on the training set. The reconstruction error threshold to separate healthy from degraded states is defined based on the classification metrics for the training set. The model thus yields a high accuracy for the training set; however, it presents a low F1-score, meaning that most degraded states are mislabeled as healthy. Results for the test set show that the model mislabels all the degraded states as healthy, and thus this model is not beneficial for diagnostics purposes. Further, implementing AE comes at a cost of training a NN, which is time consuming and adds uncertainty to the results. Time is spent trying to find the hyperparameters that best represent the data, and once this process is finished there is no guarantee that the latent variable will be able to separate degraded from healthy states.

Classification models are trained for scenarios 2 and 3. As it is shown in Table 5, these models overfit on the training set, probably caused by the class imbalance, yielding classification metrics similar to the AE from scenario 1. A fairer comparison to the proposed methodology is presented in scenario 4, where the training classes are balanced and thus the classification models learn to represent both classes equally. As it has been mentioned, not balancing the classes before training the diagnostics classifiers can also lead to intractable training times. This is particularly true for the RF and SVM. The RF models training time averaged at 657s for the unbalanced classes, while the SVM did not converge after three days training. For scenario 4, the training process was completed in under 5s for the RF models. According to Table 5, in general, the PCA processing yields better results than the AE.

Table 5. Classification metrics comparison for other preprocessing methodologies applied to the VRU system.

| Methodology | Training | | | | Test | | | |
|---|---|---|---|---|---|---|---|---|
| | Accuracy | False Healthy | False Degraded | F1 Score | Accuracy | False Healthy | False Degraded | F1 Score |
| AE (unsupervised) | 0.99 | 0.01 | 0.01 | 0.62 | 0.50 | 0.50 | 0.00 | 0.33 |
| AE-RF | 0.99 | 0.00 | 0.00 | 1.00 | 0.53 | 0.47 | 0.00 | 0.40 |
| AE-NN | 0.99 | 0.01 | 0.00 | 0.99 | 0.50 | 0.50 | 0.00 | 0.33 |
| PCA-RF | 1.00 | 0.00 | 0.00 | 1.00 | 0.57 | 0.43 | 0.00 | 0.57 |
| PCA-NN | 0.99 | 0.01 | 0.00 | 0.99 | 0.50 | 0.50 | 0.00 | 0.50 |
| AE-RF (Balanced) | 1.00 | 0.00 | 0.00 | 1.00 | 0.94 | 0.06 | 0.00 | 0.94 |
| AE-NN (Balanced) | 1.00 | 0.00 | 0.00 | 1.00 | 0.79 | 0.20 | 0.01 | 0.79 |
| PCA-RF (Balanced) | 1.00 | 0.00 | 0.00 | 1.00 | 0.98 | 0.02 | 0.00 | 0.98 |
| PCA-NN (Balanced) | 0.98 | 0.00 | 0.02 | 0.98 | 0.98 | 0.00 | 0.02 | 0.98 |

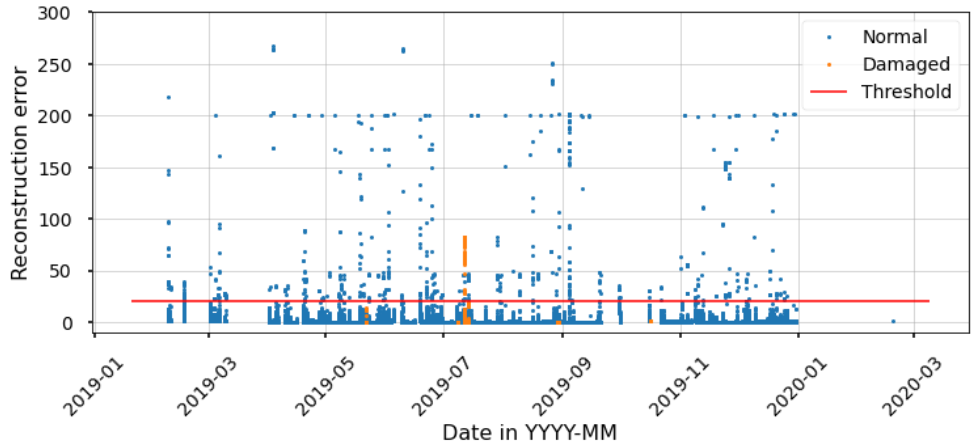

Figure 15: Autoencoder training set reconstruction error and threshold for class separation.

Although their performance is significantly improved when trained on balanced training sets, the proposed preprocessing methodology still yields better results than preprocessing the input data through PCA. This performance difference can be linked to the outlier detection process, since outlier points do not correspond to random ambient noise. Rather, as it is discussed in this Section 4.2.1, outliers are likely to come from anomalous behaviors, warm-up and cool-down periods, and external factors. As such, the AE and PCA methods are unlikely to identify these behaviors. Further, eliminating the transition state between health states is likely to cause an overlap between the classes, which could explain the higher number of falsely classified healthy states.

Considering the competitive results obtained with the preprocessing methodologies from scenario 4, it is important to highlight that these still present several disadvantages when compared to the preprocessing methodology proposed in this paper. For instance, discarding principal components with low variance can lead to information loss on the system's faulty states [70]. Therefore, blindly applying PCA for dimensionality reduction can lead to lower performance in the diagnostic models. Further, when using the latent space from the AE or the principal components from the PCA, there is an important loss of feature interpretability. There is also information loss, which is a direct consequence from the feature reduction (Figure 14). Lastly, note that to evaluate scenarios 2-4 the label generation process proposed in Section 5 needs to be used, which makes it an essential tool for the diagnostics model training process.

*6.5 Analysis and discussion*

Overall, the results for all four models presented in Table 4 are encouraging and show the potential and usefulness of the proposed preprocessing methodology when training DDMs for PHM in real systems. In this regard, the following are advantages and disadvantages of the proposed preprocessing framework.

One of the main disadvantages comes from the degraded-state assumption when defining a time-window before a failure. As it has already been discussed, the length of the time-window will be highly dependent on the system under study. There is also no guarantee that the degradation state will be present, while preventing an overlap between the healthy and degraded states is a challenging task. Nevertheless, results presented in Table 4 show that most of these drawbacks can be overcome by choosing an appropriate DDM. The overlap between healthy and degraded states can also be avoided if a larger transition state window is chosen. The proposed methodology is thought for complex mechanical systems equipped with sensor networks acquiring data in the form of time series. The preprocessing framework has not been tested for manually crafted features obtained from signal processing techniques (i.e., frequency and time-frequency domain features). It is then suggested that if signal measurements are available, these should be adapted into a time-series domain. Then, these can be treated in the same manner as any other physical sensors as recommended by the proposed approach. The proposed approach is developed with mechanical physical assets in mind and the failure mechanisms related to them. Further, this framework has not been tested or applied to electrical systems, which usually present failure mechanisms that develop much faster in comparison to mechanical degradation.

An advantage of the proposed methodology is that it can be automatized once the data set is processed once. From Figure 4 there are multiple steps that can be completely automatized, such as the label degradation generation described, the NaN feature evaluation, and the statistical analysis for feature reduction. There are two steps where manual input is required: the outlier detection and the identification of features unrelated to the system. In the case of the outlier detection, upper and lower boundaries cutoff thresholds need to be set each sensor feature. This manual process only needs to be done manually once, and it can then be automatized to process newly arrived data.

The methodology is flexible in the sense that it allows the incorporation of expert knowledge and additional information if available. For instance, nominal sensor values can be used as a preliminary outlier detection and for a more accurate healthy-labeled da-ta points. Moreover, depending on the available maintenance records, labels can be generated either for component-specific or failure mechanism, as it was showcased through both case studies. Additional steps can be added to any of the stages of the proposed methodology. For example, the health state label generation can easily be transformed into RUL labels in the case enough failure events are registered in the maintenance logs. Different normalization approaches could be used before training the DDMs as well as any of the other processing needs that might be encounter such as the one mentioned in Section 4.3. A key point to emphasize is that the only stage where expert knowledge is truly needed is in the label generation described in Section 5. Failure and operational times are essential to define the stoppage and, thus, accurate labels.

The comparison with other preprocessing methodologies shows that two of the most important steps in the proposed framework are the health state label generation and the class balance for the training set. The feature reduction is proven to be equivalent, and slightly better than a PCA, without the disadvantage of losing interpretability on the input features to the diagnostics models. Further, the PCA does not perform outlier detection, and the results presented in Table 5 suggest that the proposed preprocessing methodology can be complemented with a PCA, as it was discussed in Section 4.3.2.

## 7. Conclusions

This paper has presented a comprehensive step-by-step preprocessing methodology for sensor monitoring data of mechanical physical assets aimed towards Data-Driven models. A methodology for the generation of health state labels is also proposed for diagnostics models. Challenges faced when preparing a data set to train and test DDMs for PHM frameworks are discussed. Throughout two case studies from real systems, it is shown that data from sensor networks present common characteristics and challenges that can be addressed with the proposed preprocessing framework. A discussion was presented on how expert knowledge can have a major influence on most data preprocessing steps. As such, expert knowledge should not be neglected when processing the data.

Four different DDMs are presented to validate the proposed preprocessing methodology, showing that it is an effective and conservative approach for diagnostics models. The methodology was also compared to other preprocessing techniques such as AE and PCA for unsupervised and supervised models, where the proposed framework showed to be competitive and have several advantages over those approaches. As such, this paper takes a step towards closing the gap between the PHM models developed in academia towards their application in real systems. The presented results should encourage researchers to explore further preprocessing tools and report their preprocessing steps in detail whenever ML or DL frameworks are proposed in the context of PHM. Even though applications will tend to converge to system-specific solutions, following a transparent and explainable data processing methodology can serve to better implement and replicate data-driven diagnostics in real complex systems.

**Author Contributions:** S.C.M.: conceptualization, methodology, software, validation, formal analysis, investigation, data curation, writing—original draft preparation, visualization. E.L.D.: conceptualization, methodology, resources, supervision, writing—review & editing, project administration. M.M.: funding acquisition, resources, supervision, writing—review & editing. All authors have read and agreed to the published version of the manuscript.

**Acknowledgments:** Sergio Cofre-Martel would like to thank the Agencia Nacional de Investigación y Desarrollo (ANID – Doctorados Becas Chile - 72190097). The authors would also like to thank Camila Correa-Jullian and Mohammad Pishahang for their valuable inputs on both case studies.